\documentclass{article}

\usepackage{microtype}
\usepackage{graphicx}
\usepackage{subfigure}
\usepackage{booktabs} % for professional tables
\usepackage{hyperref}

\usepackage[accepted]{icml2023}

\usepackage{amsmath}
\usepackage{amssymb}
\usepackage{mathtools}
\usepackage{amsthm}

\usepackage[capitalize,noabbrev]{cleveref}

\theoremstyle{plain}

\theoremstyle{definition}

\theoremstyle{remark}

\usepackage{caption}

\usepackage[textsize=tiny]{todonotes}

\icmltitlerunning{Learning Dense Correspondences between Photos and Sketches}

\begin{document}

\twocolumn[
\icmltitle{Learning Dense Correspondences between Photos and Sketches}
\vspace{-0.1in}

\icmlsetsymbol{equal}{*}

\begin{icmlauthorlist}
\icmlauthor{Xuanchen Lu}{ucsd}
\icmlauthor{Xiaolong Wang}{ucsd}
\icmlauthor{Judith E. Fan}{ucsd,stanford}

\end{icmlauthorlist}

\icmlaffiliation{ucsd}{University of California, San Diego}
\icmlaffiliation{stanford}{Stanford University}

\icmlcorrespondingauthor{Judith Fan}{jefan@stanford.edu}

\icmlkeywords{Machine Learning, ICML}
\vspace{0.1in}
{%v
  \begin{center}
    \centering
    \includegraphics[width=0.92\linewidth]{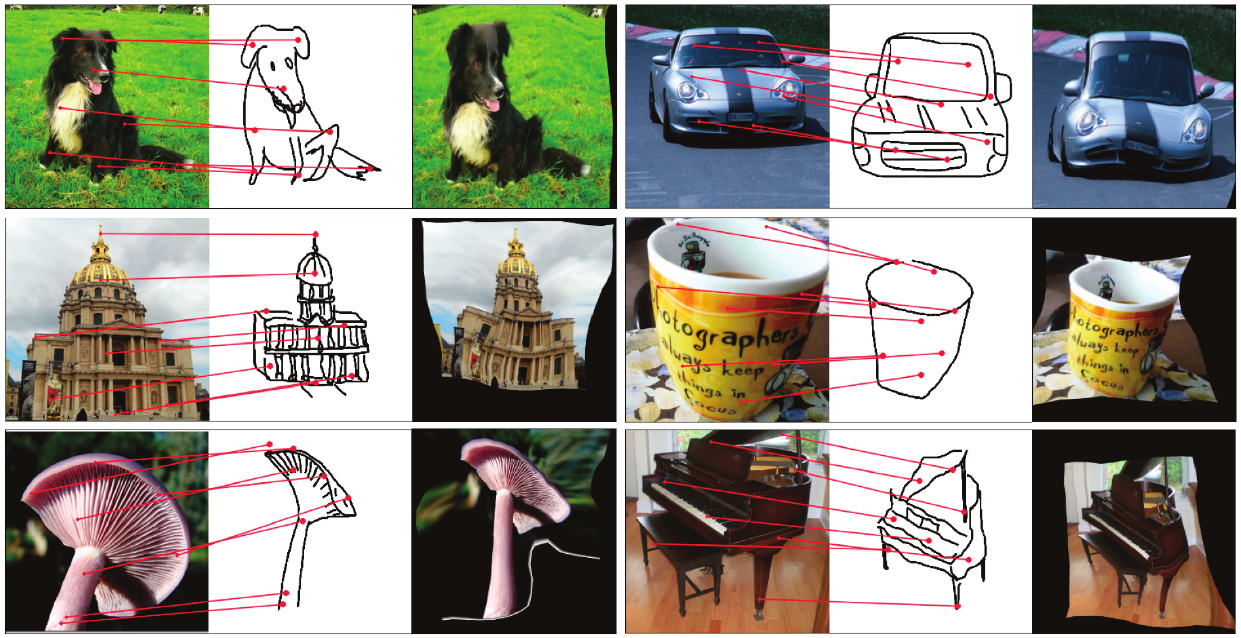}
    \captionof{figure}{
      We propose a self-supervised method for learning the dense correspondence between sketches and photos. For each photo-sketch pair, we show the annotated keypoints from our benchmark dataset \textit{PSC6K} (first column), the predicted correspondences (second column), and the result of warping the photo to the sketch (third column).
    }
    \label{fig:teaser}
  \end{center}
}
\vspace{0.1in}
]

\printAffiliationsAndNotice{}  % leave blank if no need to mention equal contribution

\begin{abstract}
Humans effortlessly grasp the connection between sketches and real-world objects, even when these sketches are far from realistic.
Moreover, human sketch understanding goes beyond categorization --- critically, it also entails understanding how individual elements within a sketch correspond to parts of the physical world it represents. 
What are the computational ingredients needed to support this ability?
Towards answering this question, we make two contributions: first, we introduce a new sketch-photo correspondence benchmark, \texttt{PSC6k}, containing 150K annotations of 6250 sketch-photo pairs across 125 object categories, augmenting the existing Sketchy dataset \cite{sangkloy2016sketchy} with fine-grained correspondence metadata.
Second, we propose a self-supervised method for learning dense correspondences between sketch-photo pairs, building upon recent advances in correspondence learning for pairs of photos.
Our model uses a spatial transformer network to estimate the warp flow between latent representations of a sketch and photo extracted by a contrastive learning-based ConvNet backbone. 
We found that this approach outperformed several strong baselines and produced predictions that were quantitatively consistent with other warp-based methods. 
However, our benchmark also revealed systematic differences between predictions of the suite of models we tested and those of humans.
Taken together, our work suggests a promising path towards developing artificial systems that achieve more human-like understanding of visual images at different levels of abstraction. Project page: {\small\url{https://photo-sketch-correspondence.github.io}}
\end{abstract}

\section{Introduction}
\label{sec:intro}

Sketching is a powerful technique humans use to create images that capture key aspects of the visual world.
It is also among the most enduring and versatile of image generation techniques, with the earliest known sketch-like images dating to at least 40,000-60,000 years ago \cite{hoffman2018dating, aubert2014pleistocene}.
Although the retinal image cast by a sketch and a real-world object are highly distinct, humans are nevertheless able to grasp the meaning of that sketch at multiple levels of abstraction, including the category label that best applies to it, the specific object instance it represents, as well as detailed correspondences between elements in the sketch and the parts of the object \cite{fan2018common,mukherjee2019communicating, yang2021visual}.
What are the computational ingredients needed to achieve such robust image understanding across domains and at multiple levels of abstraction?

\textbf{Generalizing across photorealistic and stylized image distributions.}
There has been substantial recent progress in the development of artificial vision systems that capture some key aspects of sketch understanding, especially sketch categorization and sketch-based image retrieval \cite{eitz2012sketch,sangkloy2016sketchy, yu2016sketch, yu2017sketch, bhunia2020sketch}. 
In addition, the availability of larger models that have been trained on vast quantities of paired image and text data have led to encouraging results on tasks involving images exhibiting different visual styles \cite{radford2021learning}, including sketch generation \cite{vinker2022clipasso}.
However, recent evidence suggests that even otherwise high-performing vision models trained on photorealistic image data do not generalize well to other image distributions as well as neurons in primate inferotemporal cortex (a key brain region supporting object categorization) \cite{bagus2022primate}, indicating that a large gap remains between the capabilities of current computer vision systems and those achieved by biological systems.

\textbf{Perceiving semantic correspondences between images.} 
In particular, a core open problem in human sketch understanding concerns the computational ingredients required to encode the internal structure of a sketch with sufficient fidelity to establish a detailed mapping between parts of a sketch with parts of the object it represents \cite{kulvicki2015analog, fodor2007revenge}. 
The problem of discovering semantic correspondences between images is a well established problem in computer vision. 
In the typical setting, the goal is to establish dense correspondences between images containing objects belonging to the same class.
Classical methods \cite{berg2005shape, kim2013deformable, liu2010sift} determine the alignment with hand-crafted feature descriptors such as SIFT \cite{lowe1999object} or DOG \cite{dalal2005histograms}.
More recently developed methods \cite{ham2016proposal, rocco2018a, WarpC}, which benefit from the robust feature representations learned by deep neural networks are more robust to variations in appearance and shape. 
However, finding correspondence between photos and sketches is particularly challenging as human-generated sketches are inherently selective, highlighting the most relevant aspects of an object's appearance at the expense of other aspects \cite{fan2020pragmatic, huey2021explanatory}.
Moreover, sketches typically lack the texture and color cues that can facilitate dense correspondence learning for color photos.
As a consequence, the task of learning dense semantic correspondences between photos and sketches relies on a substantial degree of visual abstraction in order to establish strong semantic alignment between images from different modalities. 

\textbf{Self-supervised representation learning.}
A robust finding from the past decade is that deep neural networks trained with supervision at large, labeled image datasets can achieve state-of-the-art performance \cite{krizhevsky2017imagenet,simonyan2014very, he2016deep}. 
Moreover, models trained in this way currently provide the most quantitatively accurate models of biological vision in non-human primates and humans \cite{yamins2014performance, khaligh2014deep, rajalingham2018large,cadena2019deep}.
Nevertheless, such models are unlikely to explain how humans are capable of achieving such robust image understanding across different modalities given the implausibility that such large, labeled datasets were available to or necessary for humans to learn to understand natural visual inputs, much less to interpret sketches \cite{hochberg1962pictorial,kennedy1975outline}. 
Recent advances in self-supervised representation learning have begun to approach the performance of supervised models without the need for such labels \cite{ wu2018unsupervised, he2020momentum}, while also emulating key aspects of visual processing in biological systems \cite{zhuang2021unsupervised, konkle2020instance}.
However, it remains unclear to what degree these advances are sufficient to support challenging multi-domain image understanding tasks, including predicting dense photo-sketch correspondences.

\textbf{Our contributions: Evaluating a self-supervised method for learning photo-sketch correspondences.} 
Towards meeting these challenges, our paper makes two key contributions: first, we establish a new benchmark for photo-sketch dense correspondence learning: \texttt{PSC6k}.
This benchmark consists of 150,000 pairs of keypoint annotations for 6250 photo-sketch pairs spanning 125 object categories. 
Each annotation consists of a keypoint marked by a human participant on an object in a color photo that they judged to correspond to a given keypoint appearing on a sketch of the same object.
All photo-sketch pairs were sampled from the well established Sketchy dataset \cite{sangkloy2016sketchy}, a collection of ~75K sketches produced by humans to depict objects in 12.5K color photographs of objects spanning 125 categories. 

Our second contribution is a self-supervised method for learning photo-sketch correspondences that leverages a learned nonlinear ``warping'' function to map one image to the other. 
This approach embodies the hypothesis that sketches preserve key information about spatial relations between an object's constituent parts, even if they also manifest distortions in the size and shape of these parts.
This hypothesis is motivated by the view that representational line drawings, as sparse as they are, are meant to accurately convey 3D shape \cite{hertzmann2020line}, which stands in sharp contrast to the view that the relationship between drawings and objects are established purely by convention \cite{goodman1976languages}.
Nevertheless, the nonlinear ``warping'' approach we propose diverges from very strong versions of the 3D-shape-preservation account \cite{greenberg2021semantics}, which are not well equipped to handle the kinds of nonlinear visual distortions that human-generated sketches exhibit \cite{eitz2012sketch, sangkloy2016sketchy, fan2018common}.

Our system consists of two main components: the first is a multimodal image encoder trained with a contrastive loss \cite{wu2018unsupervised,zhuang2021unsupervised}, with photos and sketches of the same object being treated as positive examples, and those depicting different objects as negative examples. 
The second component is a spatial transformer network \cite{jaderberg2015spatial} that estimates the transformation between each photo and sketch and aims to maximize the similarity between the feature maps for both images.
Using our newly developed \texttt{PSC6k} benchmark, we find that our system outperforms other existing self-supervised and weakly supervised correspondence learning methods, and thus establishes the new state-of-the-art for sketch-photo dense correspondence prediction.  
We will publicly release \texttt{PSC6k} with extensive documentation and code to enhance its usability to the research community.

\begin{figure}[htp]
    \centering
    \includegraphics[width=0.43\textwidth, trim=8 8 8 8, clip]{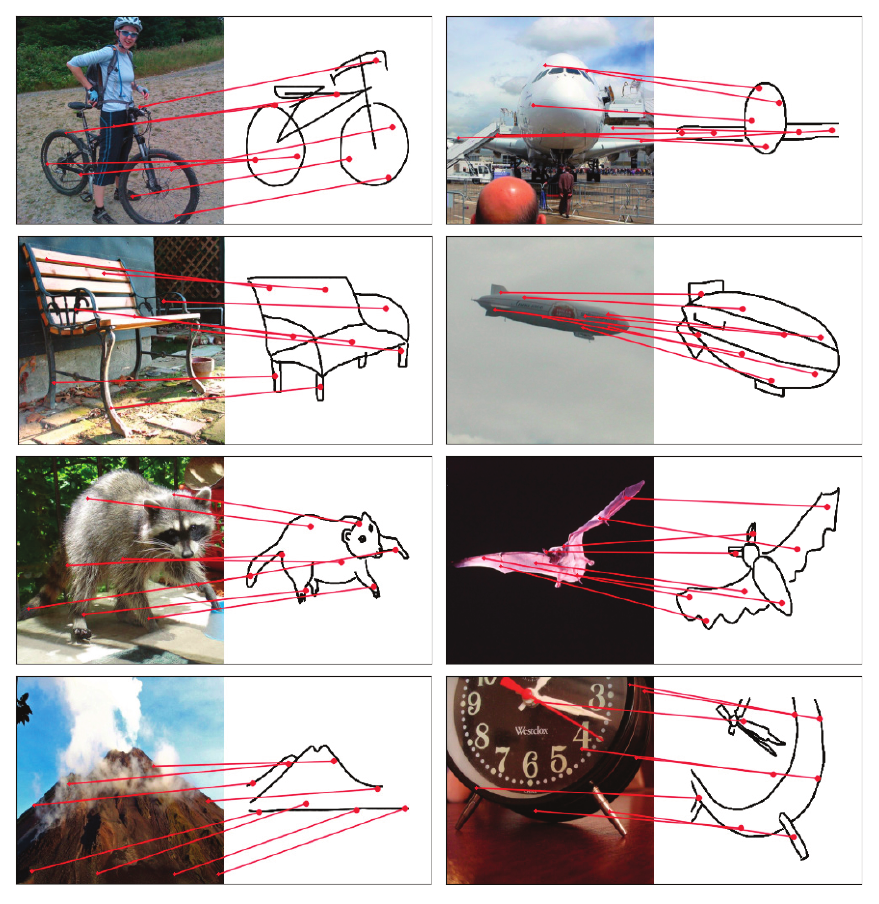}
    \caption{Examples of human-annotated photo-sketch pairs from our new photo-sketch correspondence benchmark \texttt{PSC6k}.}
    \label{fig:dataset}
    \vspace{-6mm}
\end{figure}

\section{Photo-Sketch Correspondence Benchmark (\texttt{PSC6k})}
\label{sec:benchmark}

Our first goal was to establish a novel photo-sketch correspondence benchmark satisfying two criteria: first, it should build directly upon existing benchmarks in sketch understanding and second, it should provide broad coverage of a wide variety of visual concepts. 
Towards that end, we developed \texttt{PSC6k} by directly augmenting the Sketchy dataset \cite{sangkloy2016sketchy}, which already contains 75,471 human sketches produced from 12,500 unique photographs spanning 125 object categories. 

\subsection{Sampling Photo-Sketch Pairs}
We sampled photo-sketch pairs from the original test split of the Sketchy dataset, which consisted of 1250 photos and their corresponding sketches. 
We manually filtered out sketches that were completely off-target or that depicted the photographed object from the wrong perspective \cite{sangkloy2016sketchy}. 
We then randomly sampled 5 sketches from among the remaining valid sketches produced of each photo, resulting in 6250 unique photo-sketch pairs.

\begin{figure*}[htp]
    \centering
    \includegraphics[width=1.0\textwidth, trim=2 2 2 2, clip]{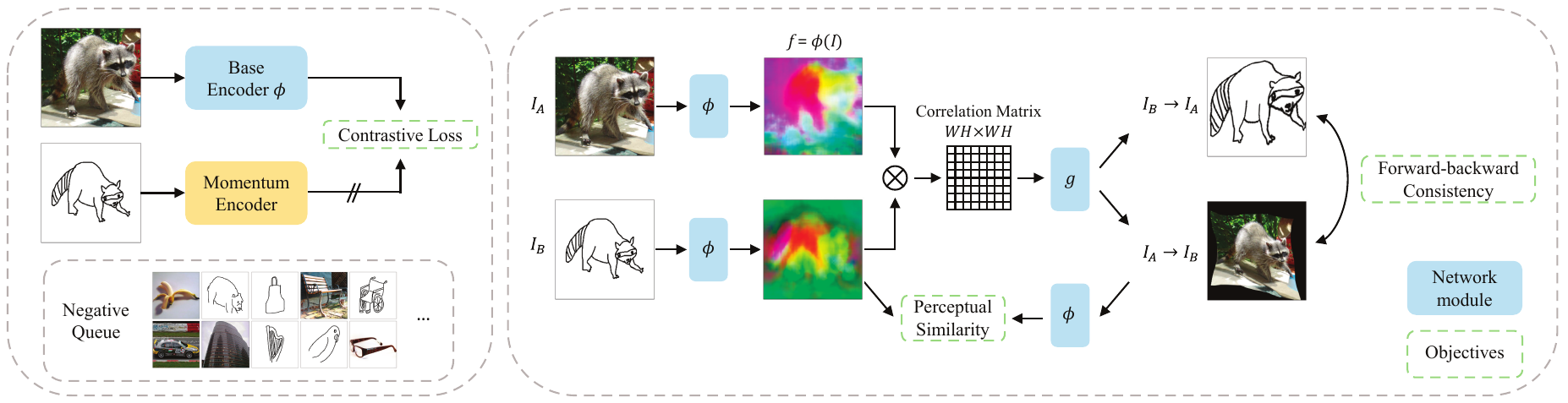}
    \vspace{-4mm}
    \caption{We propose a self-supervised framework for learning photo-sketch correspondence by estimating a dense displacement field that warps one image to the other. The framework consists of a multi-modal feature encoder that aligns the photo-sketch representation with a contrastive loss, and an STN-based warp estimator to predict transformation that maximizes the similarity between feature maps of the two images. The estimator learns to optimize a combination of weighted perceptual similarity and forward-backward consistency.}
    \label{fig:diagram}
\end{figure*}

\subsection{Collecting Human Keypoint Annotations}
We formalize the problem of identifying photo-sketch correspondences as the ability to map a keypoint located on a sketch to the location in the source photograph that best corresponds to it. 
For example, a keypoint appearing on the left wing of a sketch of an airplane should be mapped to the ``same'' location on the left wing of the photograph of that same airplane.  
For each photo-sketch pair, we sampled 8 keypoints spanning as much of the object as possible. 
To determine these keypoints, we first computed segmentation masks for each sketch, relying upon the heuristic that outermost contour of the sketch naturally serves as the contour of the object in the sketch. 
The pixels covered by the segmentation mask were then clustered into 8 groups to estimate 8 ``pseudo-part'' regions. We employ nearest-neighbor-based spectral clustering to prioritize connectivity within each pseudo-part. 
A keypoint was then placed at the centroid of each pseudo-part.

This approach allowed us to automatically discover regions of the sketch that are likely to be semantically meaningful without the need for explicit part labels.
However, this approach is also less sensitive to sketch regions that constitute only a small portion of object mask (e.g., a cat's whiskers). 
As such, future work could employ a combination of region-based and stroke-based keypoints to gain fuller coverage of semantically meaningful regions of sketches.

Next, we recruited 1,384 participants using the Prolific crowdsourcing platform to provide annotations.
Participants provided informed consent in accordance with the UC San Diego Institutional Review Board (IRB).
On each trial, participants were cued with a keypoint appearing on a sketch and asked to indicate its corresponding location in a photo appearing next to it. 
Each participant provided annotations for 125 photo-sketch pairs, one from each category. 
We collected three annotations from different participants for each keypoint in every sketch, resulting in 150,000 annotations across all 6250 photo-sketch pairs.
We defined the centroid over these annotations as the ground-truth keypoint in the photo.
In rare cases, there was one annotation out of three with an exceptionally large distance from the median location of all three annotations; these responses were flagged as outliers and excluded from the determination of the centroid. 
See \autoref{sec:supp benchmark} for additional details regarding the creation of this photo-sketch correspondence benchmark. 

\section{Weakly-supervised Photo-Sketch Correspondence}
\label{sec:method}
In this section, we present our weakly-supervised model for finding the pixel-level correspondence between photo-sketch pairs. We formulate the problem as estimating the displacement field across a sketch $I_s\in \mathbb{R}^{h\times w \times 3}$ and a photo $I_p\in \mathbb{R}^{h\times w \times 3}$ that depict the same object (\autoref{fig:diagram}). 
Our goal is to find the cross-modal photo-sketch alignment in a weakly-supervised manner, by maximizing the perceptual similarity of an image in $(I_p, I_s)$ and its warped counterpart. Our framework consists of a feature encoder $\phi$ that learns a shared feature space of photo and sketch, and a warp estimator $T$ based on the spatial transformer network (STN) that directly predicts the displacement field $F \in \mathbb{R}^{h\times w \times 2}$, where we extract the dense correspondence. 

\subsection{Feature Encoder $\phi$}
Here we leverage advances in contrastive learning to develop a weakly-supervised feature encoder on photo-sketch data pairs. Contrastive learning obtains a feature representation by contrasting similar and dissimilar pairs. 
Here, the photo $I_p$ and the sketch $I_s$ depicting the same object become a natural choice to construct similar pairs. 
Unlike typical contrastive learning schemes \cite{wu2018unsupervised, chen2020simple, he2020momentum} that take augmented views of the same image $I$ as positives, our model uses augmented views from the same photo-sketch pair $(I_p, I_s)$. 
To minimize the contrastive loss over a set of photo-sketch pairs, the encoder must learn a feature space that attracts photo/sketch from the same pair and separates photo/sketch from distinct pairs. 

Similar to \cite{he2020momentum}, we formulate pair-level contrastive learning as a dictionary look-up problem. For a given photo-sketch pair $(I_p, I_s)$, random data augmentation is applied to generate the view pair $(\widetilde{I_p}, \widetilde{I_s})$. One view in the pair is randomly selected as the query and the other becomes the corresponding key. We denote their representations encoded by $\phi$ as $q$ and $k^+$, respectively. The query token $q$ should match its key $k^+$ over a set of negative keys ${k^-}$ sampled from other photo-sketch pairs. To optimize this target, we minimize InfoNCE \cite{oord2018representation} as follows:
    \begin{equation}\label{eq:infonce}
        \mathcal{L}_{nce} = -\log{\frac{\exp\left({q{\cdot}k^{+} / \tau}\right)}{\exp{\left({q{\cdot}k^{+} / \tau}\right)} + \sum_{k^{-}}{\exp\left({q{\cdot}k^{-} / \tau}\right)} }},
    \end{equation}
where $\tau$ is a temperature hyperparameter scaling the data distribution in the metric space. 

To explore the inherent similarity between photos and sketches, we use a shared encoder $\phi$ for images from both modalities. We replace the batch normalization (BN) \cite{ioffe2015batch} in the encoder with conditional batch normalization \cite{de2017modulating} for better domain alignment. Detailed implementation and experiment are reported in \autoref{sec:result}.

\subsection{Warp Estimator $T$}

Given the source and target image $I_s, I_t$ and their representation $X_s, X_t$, the warp estimator $T$ predicts the displacement field $F_{I_s\to I_t} = T(X_s, X_t)$. Inspired by \cite{sun2018pwc}, we propose a simplified pyramidal warp estimation module for the ResNet backbone.

{\bf Affinity function $f$.}
While it is possible to estimate the correspondence based on the feature affinity at a specific layer of the encoder $\phi$, e.g., the final convolutional layer, it is beneficial to evaluate affinities at multiple layers along the feature pyramid. We select a set of $n$ feature layers of interest, denoted as $X_s = \{x^i_s\}_{i=0}^{n-1}$ and $X_t=\{x^i_t\}_{i=0}^{n-1}$. We bilinearly upsample all selected feature maps to the same spatial resolution, and concatenate them along the channel dimension for the multi-layer feature maps, $X_s \in \mathbb{R}^{c \times h \times w}$ and $X_t \in \mathbb{R}^{c \times h \times w}$.

With the source and target feature maps $X_s$ and $X_t$, we compute affinity as the correlation between feature embeddings: with pixel $i$ in feature map $X_s$ and pixel $j$ in feature map $X_t$, $A_{(s, t)}(i, j) = X_s(i)^T X_t(j)$. 
The pairwise affinity between every pixel in the source and target feature maps forms the affinity matrix $f(X_s, X_t):= A_{(s,t)} \in \mathbb{R}^{hw \times hw}$.  

\begin{figure}[ht]
    \centering
    \includegraphics[width=0.40\textwidth, trim=4 4 4 2, clip]{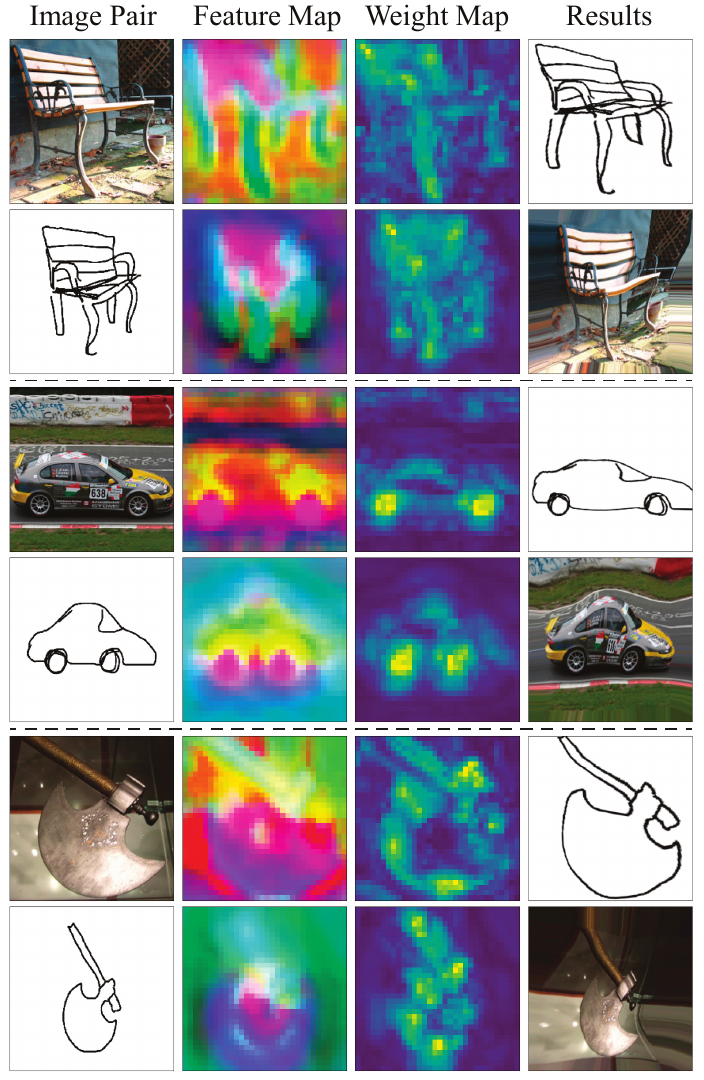}
    \caption{Example image pairs, feature maps, weight maps, and final results processed in our warp estimator. The weight maps highlight semantic parts that have the largest correlation between the two images. We use PCA to project the feature dimensions to 3 principal components as RGB.}
    \label{fig:warp estimator}
    \vspace{-4mm}
\end{figure}

{\bf Estimation Module $g$.}
Module $g$ takes the affinity matrix $A_{(s,t)}$ and directly estimates the displacement field $F$ from the source image to the target image. Following the idea of coarse-to-fine refinement, it consists of three STN blocks at different scales with residual connections, denoted as $g_1$, $g_2$ and $g_3$. Each STN-block (except the first block) takes the affinity matrix warped by the previous block and regresses a new displacement field to refine the alignment. The first block $g_1$ regresses at the $4 \times 4$ scale, estimating displacement field $F^{(0)} \in \mathbb{R}^{4\times 4 \times 2}$. $g_2$ and $g_3$ regress at the $8 \times 8$ and $16 \times 16$ scale, respectively. The displacement field at each block is computed as
    \begin{gather}\label{eq:f0}
        F^{(1)} = g_1(f(X_s, X_t)),\\
        F^{(k)} = F^{(k-1)} + g_i(f(warp(X_s, F^{(k-1)}), X_t)),
    \end{gather}
where $warp(I, F)$ operation warps image $I$ to target according to the displacement field $F$. It is implemented with bilinear interpolation.

After $g_3$ generates the $16 \times 16$ displacement field, it is upsampled to full image resolution as the final estimation.

\subsection{Weighted Perceptual Similarity}
We propose using weighted perceptual similarity to evaluate the quality of estimated displacement field between the photo-sketch pair. Instead of directly evaluating similarity using the warped source feature map (direct similarity), we pass the warped source image into the feature encoder {\it again} and evaluate similarity using the new feature map, so that the feature encoder serves as a soft constraint that reduces warping artifacts and stabilizes training (perceptual similarity). 
We use subscripts to indicate the direction of warp; for example, the displacement field from $I_s$ to $I_t$ is denoted $F_{s\to t}$. We also denote the warped image as $I_{s \to t} = warp(I_s, F_{s\to t})$. 

{\bf Perceptual similarity $s$.}
For an image pair $(I_s, I_t)$, the model estimates the flow $F_{s\to t}$ and renders the warped source image $I_{s\to t}$. The warped source image is passed through the encoder $\phi$ to generate its new set of feature maps $X_{s\to t}$, as well as its new affinity with the target $A_{(s\to t,t)}$. The new affinity matrix represents how well the warped source image semantically aligns with the target. 

In the ideal case, each pixel in the warped source $X_{s\to t}$ will have the highest correlation with the pixel at the same location in the target $X_t$. This is reflected in the affinity space $A_{(s\to t,t)}\in \mathbb{R}^{n \times hw \times hw}$ as a maximized diagonal along the second and third axes. For a pixel in warped source $X_{s\to t}$, we formulate the optimization as selecting the pixel that matches correctly from all pixels in target $X_t$:
    \begin{equation}\label{eq:similarity}
        s(n, i) = -\log{\frac{\exp\left({A_{(s\to t,t)}(n, i, i)}/ \tau\right)}{\sum_{j}{\exp\left({A_{(s\to t,t)}(n, i, j)}/ \tau\right)} }},
    \end{equation}
where $n$ is the index of the feature layer to evaluate on; $i, j$ are indices of pixels in the source and target feature map.

{\bf Weight function $w$.}
While it is possible to optimize flow estimation with the above formula, there are two problems. First, sketches contain a large number of empty pixels, and photos often suffer from background clutter. Moreover, while the encoder activation generally lies over the entire object in the photo, the activation concentrates along the strokes in a sketch. As a result, optimizing the correspondence of every pixel is inefficient and biased toward the background. To focus optimization on important matches, we consider an intuitive rule: important pixels in one image should have greater affinities to the other image. It is formulated as a weight function:
    \begin{equation}\label{eq:weight function}
        w(n, i) = scale(\max_j{[norm(A_{(s\to t,t)})(n, i)]})
    \end{equation}
where $norm$ is the normalization over the affinity matrix to penalize pixels that have multiple large affinities in the other image. $scale$ is an arbitrary operation to standardize the weight function. We use Min-Max to scale its distribution to $[0, 1]$.

Therefore, the final perceptual similarity loss is given by 
    \begin{equation}\label{eq:weighted similarity}
        \mathcal{L}_{sim}(n, i) = w(n, i)s(n, i) 
    \end{equation}
    
In \autoref{fig:warp estimator}, we visualize the image pairs, feature maps, weight maps, and the final alignment results of photo-sketch pairs from \textit{PSC6k} to exhibit the function of each component in our estimator.

\begin{table*}[t]
\centering
\begin{tabular}{ll|c@{~~}c|c@{~~}c}
 &  & \multicolumn{2}{c}{\textbf{Transfer}} & \multicolumn{2}{c}{\textbf{Retrain}} \\
Methods  & Encoder & PCK-5 & PCK-10 & PCK-5 & PCK-10\\ \midrule
CNNGeo~\cite{rocco2018a} & ResNet-101 & 27.59 & 57.71 & 19.19 & 42.57\\ 
WeakAlign~\cite{rocco2018a} & ResNet-101 & 35.65 & 68.76 & 43.55 & 78.60\\
NC-Net~\cite{rocco2018b} &  ResNet-101  & 40.60 & 63.50 & --& -- \\
DCCNet~\cite{DCCNet}  & ResNet-101 & 42.43 & 66.53 & -- &  -- \\
PMD~\cite{PMD} & VGG-16 & 35.77 & 71.24 & -- & -- \\ 
WarpC-SemanticGLUNet~\cite{WarpC} & VGG-16 & 48.79 & 71.43 & 56.78 & 79.70 \\ 
\textbf{Ours} & ResNet-18 & -- & -- & 56.38 & \textbf{83.22}\\ 
\textbf{Ours} & ResNet-101 & -- & -- & \textbf{58.00} & \textbf{84.93}\\ 
\bottomrule 
\end{tabular}
\caption{State-of-the-art comparison for photo-sketch correspondence learning.}
\label{tab:comparison}
\end{table*}

\subsection{Additional Objectives}
 In addition to the perceptual similarity loss, we consider an additional self-supervised loss to assist robust warp estimation and stabilize training.  

{\bf Forward-backward consistency.} Forward-backward consistency is a classical idea in tracking \cite{vondrick2018tracking, wang2019learning, jabri2020space} and flow estimation \cite{meister2018unflow, rocco2017convolutional, jeon2018parn, WarpC, DCCNet} as constraints. Namely, we expect the estimated forward flow $F_{s\to t}$ to be the inverse of the estimated backward flow $F_{t\to s}$. It poses a strict constraint on the network for symmetric prediction. We minimize the $L2$ norm between the identity flow and the composition of the forward flow and backward flow:
    \begin{equation}\label{eq:consistency}
        \mathcal{L}_{con} = \|warp(F_{s\to t}, F_{t\to s}) - F_{\mathbb{I}}\|,
    \end{equation}
where $F_{\mathbb{I}}$ is the identity displacement that maps all locations to themselves.

Overall, our final objective is
    \begin{equation}\label{eq:final}
        \mathcal{L} = \lambda_{sim}\mathcal{L}_{sim} + \lambda_{con}\mathcal{L}_{con} ,
    \end{equation}

\section{Experiments}
\label{sec:result}
Here we empirically evaluate our method and compare it to existing approaches in dense correspondence learning on the photo-sketch correspondence benchmark. We analyze the difference between human annotations and predictions from existing methods. We show that our method establishes the state-of-the-art in the photo-sketch correspondence benchmark and learns a more human-like representation from the photo-sketch contrastive learning objectives.
We conducted additional experiments to evaluate generalization to unseen categories in \autoref{sec:supp unseen}.

\subsection{Implementation Details}
The input image size is set to 256 following our photo-sketch correspondence benchmark. We use ResNet-18 and ResNet-101 as our feature encoder. The encoder is initialized with pretrained weights from MoCo training \cite{he2020momentum} on ImageNet-2012 \cite{imagenet}. We then train our encoder on the training split of Sketchy for 1300 epochs. Since there are multiple sketches for each photo in the dataset, at each epoch, we iterate through all photos and sample a corresponding sketch for each photo.
We follow the recipe from MoCo \cite{he2020momentum, chen2020improved}, with $dim=128, m=0.999, t=0.07, lr=0.03$ and a two-layer MLP head. Noticeably, we set the size of the memory queue to $K=8192$ to prevent multiple positive pairs from appearing at the same time. 

We then train the estimator for 1200 epochs with a learning rate of $0.003$, leading to 2500 epochs of training in total. We set the weights of the objectives to $\lambda_{sim}=0.1, \lambda_{con}=1.0$. We compute $\mathcal{L}_{sim}$ using the features after ResNet stages 2 and 3, and the temperature is set to $\tau = 0.001$.

We apply the same set of augmentations to both feature encoder and the warp estimator, consisting of random color jitter, grayscale, and Gaussian blur, which are consistent with the settings in MoCo v2 \cite{chen2020improved} and SimCLR \cite{chen2020simple}. However, we replace random cropping with a combination of affine and TPS transformations for a more complex spatial distortion.

We train the network with the SGD optimizer, a weight decay of $1e-4$, a batch size of $256$, and the native mixed precision from PyTorch. We adopt a cosine learning rate decay schedule \cite{loshchilov2016sgdr}.

\subsection{Photo-sketch Correspondence Estimation}
We evaluate our correspondence estimation results qualitatively and quantitatively. We compare our method with existing approaches in correspondence learning with image or pair-level supervision, and present a state-of-the-art comparison on photo-sketch correspondence in \autoref{tab:comparison}. For fair comparisons, we retrain existing open-sourced methods on the same photo-sketch dataset we used to develop our own model \cite{sangkloy2016sketchy}. We report their PCK for $\alpha = (0.05, 0.1)$ in two settings: transfer (directly evaluate on photo-sketch correspondence with pretrained weights) and retrain (train from scratch on photo-sketch correspondence). Methods that fail to converge on photo-sketch dataset are left blank. In \autoref{sec:supp eval}, we include methods with stronger supervision to the table and detail the training/evaluation setting of each method.

Our approach sets a new state-of-the-art for photo-sketch correspondence. Although we only regress flow at $16\times 16$, which is less than the granularity of PCK-05, our ResNet-101 model gains a substantial increase of +1.22\%/+5.23\% compared to the second-best method WarpC-SemanticGLU-Net \cite{WarpC}. This is surprising as the latter method benefits from flow resolution four times as large as ours, and additional two-stage training on CityScape \cite{Cordts2016Cityscapes}, DPED \cite{ignatov2017dslr}, and ADE \cite{zhou2019semantic}. Our smaller ResNet-18 model also outperforms most existing methods despite a significantly shallower feature encoder, demonstrating the effectiveness of our pair-based contrastive learning scheme in finding dense correspondences between images from different image modalities. 
We visualize more examples of the dense correspondence that our model predicts in \autoref{sec: supp qualitative}.

\subsection{Ablation Study}
We conduct two sets of ablation experiments on the ResNet-18 version of our framework. In \autoref{tab:encoder ablation} we analyze different training schemes for the feature encoder. In the first row, we directly use the pretrained weights from ImageNet contrastive learning. The following rows compare the performance of different ways of constructing positive pairs: 1) two augmented views from single images from the photo-sketch dataset, as in classical contrastive learning; 2) a photo and a sketch randomly sampled from the same class; and 3) a photo and a sketch from the same photo-sketch pair. We find that the pretrained model on ImageNet leads to the worst performance due to its failure to generalize to sketch data. Classical contrastive learning on the photo-sketch dataset also harms model estimation, because the domains of photo and sketch are not explicitly aligned in the representation space. The best result comes from contrastive learning on photo-sketch pairs, as it provides the strongest supervision for learning discriminative features. 
In \autoref{tab:full ablation}, we analyze the key components of our correspondence estimation framework. We first show the importance of our objectives, by ablating the similarity loss, the perceptual version of the similarity loss, and the consistency loss. In addition, we show that the use of the weight function, multiple feature layers, and conditional BN further improves the model performance. 

\begin{table}[t]
\centering
\resizebox{0.4\textwidth}{!}{
\begin{tabular}{l|c@{~~}c}

Training Description & PCK-5 & PCK-10 \\ 
\midrule
ImageNet only & 17.20 & 48.92 \\
CL on individual image & 44.24 & 75.84 \\
CL on image class & 54.73 & 81.78\\
CL on image pair & 56.38 & 83.22 \\
\bottomrule
\end{tabular}
}

\vspace{-1mm}
\caption{Ablation study on training feature encoder.}
\vspace{-1mm}

\label{tab:encoder ablation}
\end{table}

\begin{table}[t]
\small
\centering
\resizebox{0.4\textwidth}{!}{
\begin{tabular}{l|c@{~~}c}
Ablation Description & PCK-5 & PCK-10 \\ \midrule
No $\mathcal{L}_{sim}$ & 17.46 & 49.43 \\
No perceptual $\mathcal{L}_{sim}$ & 49.31 & 80.61 \\
No $\mathcal{L}_{con}$ & 52.25 & 80.38 \\
No weight function $w$ & 54.11 & 82.51\\
No multiple feature layers & 55.20 & 83.28 \\
No conditional BN & 55.84 & 82.67\\
Complete model & 56.38 & 83.22 \\
\bottomrule
\end{tabular}
}
\vspace{-1mm}
\caption{Ablation study on correspondence estimation.}
\vspace{-3mm}
\label{tab:full ablation}
\end{table}

\subsection{Comparing model and human error patterns}
\begin{figure}[htp]
    \hspace*{-3mm}
    \centering
    \includegraphics[width=0.5\textwidth]{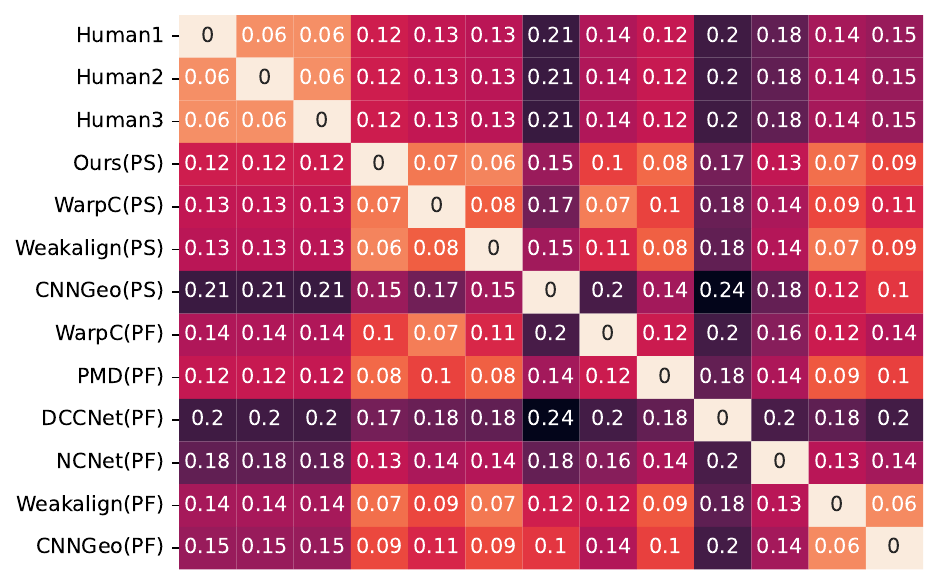}
    \vspace{-4mm}
    \caption{Measuring human and model consistency. Each cell represents the mean pixel distance between correspondence predictions generated by two systems (whether artificial or human), normalized by the image size. We denote models trained on Photo-sketch pairs with PS, and models trained on PF-Pascal \cite{ham2016proposal} as PF.}
    \label{fig:model consistency}
    \vspace{-2mm}
\end{figure}

To what degree do any of the models tested generate predictions that achieve the degree of consistency that we observe between individual human annotators?
To evaluate this question, for each pair of systems (whether two models, two humans, or a model and a human), we computed the normalized mean pixel distance between the predictions they generated for a given photo-sketch pair, then normalized this distance by the image size.
We find that while higher-performing models tend to produce predictions that are more similar to one another, all of the models taken together display systematic biases that are distinct from those of humans performing the photo-sketch correspondence task \autoref{fig:model consistency}.
These results indicate the size of the current human-model gap and suggest that future progress on this benchmark will entail bringing human-model consistency values closer to that observed between individual humans.

\subsection{Shape Bias in Learned Representation}
\begin{figure}[htp]
    \centering
    \includegraphics[width=0.48\textwidth, trim=6 0 0 0, clip]{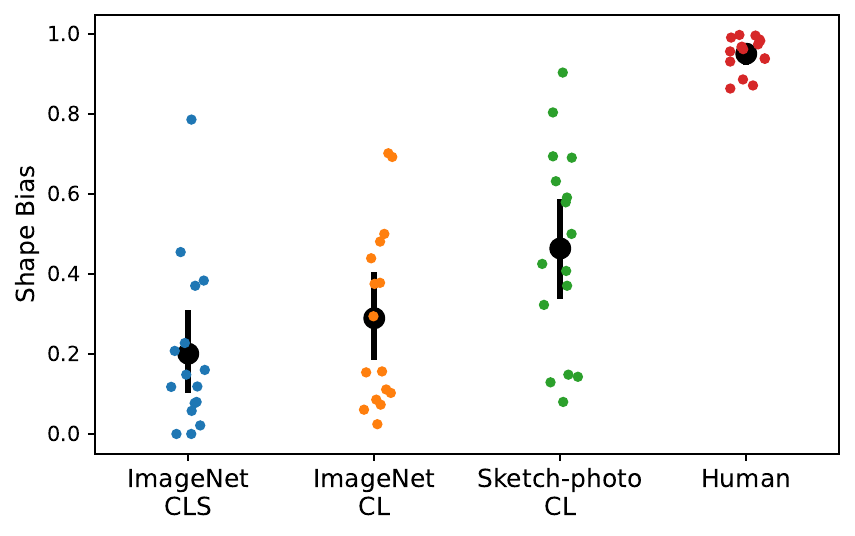}
    \vspace{-2mm}
    \caption{Comparing the degree of shape vs. texture bias between models trained with different objectives. Higher values suggest that the model recognition depends more on shape information. Our model exhibits more human-like performance. Each dot represents an object category from \cite{geirhos2018imagenet}. Error bars indicate 95\% CI.}
    \label{fig:shape bias}
    \vspace{-2mm}
\end{figure}

Recent work has shown that ImageNet-trained CNNs are biased towards object texture compared to global object shape on image recognition tasks \cite{geirhos2018imagenet}. 
Since sketch recognition requires relies on cues to object category apart from texture, we hypothesized that our photo-sketch contrastive learning pre-training procedure would mitigate this texture bias. 
To evaluate this hypothesis, we followed the same evaluation protocol as in \cite{geirhos2018imagenet, geirhos2021partial}. 
It devises a cue-conflict experiment in which a model aims to classify images with conflicting shape and texture. 
We report the shape bias of ResNet-18 models from several different training objectives: ImageNet classification (20.06\%), ImageNet contrastive learning (28.93\%), photo-sketch contrastive learning (46.36\%), and the result of human participants (95.04\%). 
The model trained on photo-sketch contrastive learning exhibits a reliably weaker texture bias (i.e., and thus stronger shape bias) than its photo-only counterparts (\autoref{fig:shape bias}). 

\section{Related Work}
\label{sec:related work}
\textbf{Self-supervised Representation Learning.} 
Learning with self-supervision aims to obtain generic representations for diverse downstream tasks with minimal dependence on human labels \cite{wang2015unsupervised, doersch2015unsupervised, pathak2016context, noroozi2016unsupervised, zhang2016colorful, gidaris2018unsupervised, wu2018unsupervised}. Recent research on sketch understanding also benefits from such development \cite{pang2020solving, xu2020deep, bhunia2021vectorization}.
These approaches are especially important for making progress towards human-like image understanding, given that large numbers of labeled images are neither available to nor necessary for humans to develop robust perceptual abilities \cite{zhuang2021unsupervised, konkle2020instance,rajalingham2018large}, including the ability to understand sketches \cite{hochberg1962pictorial, kennedy1975outline}.
In particular, recently proposed \textit{contrastive learning} techniques demonstrate competitive performance with supervised baselines not only on visual recognition \cite{hjelm2018learning, oord2018representation, wu2018unsupervised, chen2020simple, he2020momentum, grill2020bootstrap, chen2020big, chen2021empirical}, but also on learning visual representations from inputs varying across sensory views \cite{tian2020contrastive, tian2020makes}, across frames in video \cite{jabri2020space, xu2021rethinking,zhuang2020unsupervised}, and even between text and images \cite{radford2021learning, jia2021scaling}.
Here, we leverage contrastive learning-based pretraining to achieve strong performance on visual correspondence between images from highly distinct distributions (i.e., photos and sketches). 
To the best of our knowledge, ours is the first paper to successfully apply these approaches to the problem of photo-sketch dense correspondence prediction.  

\textbf{Weakly-supervised Semantic Correspondence Learning.} 
Geometric matching \cite{melekhov2019dgc, li2020dual, rocco2020efficient, shen2020ransac, truong2020glu} is perhaps the most basic form of correspondence prediction, which aims to align two views of the same scene.
On the contrary, \textit{semantic matching} \cite{ham2016proposal, rocco2018a, rocco2018b, DCCNet, PMD, WarpC} aims to establish more abstract correspondences between the image of objects in the same class, in a way that is tolerant to greater variation in appearance and shape. 
Due to difficulties in collecting ground-truth data for dense correspondence learning, prior work has generally resorted to weak supervision, such as synthetic transformation on single images \cite{rocco2018a, jeon2018parn, seo2018attentive} and image pairs \cite{rocco2018b, kim2019semantic, kim2018recurrent, jeon2020guided, DCCNet, PMD, WarpC, pwarpc}. 
Various objectives have been proposed to explore the correspondence from weak supervision, including synthetic supervision, optimization of the cost volume, forward-backward consistency, or a combination of these objectives. 
Most work utilizes hierarchical features in deep models from supervised pretraining on ImageNet. 
The dense correspondence is then predicted with a dense flow field \cite{ham2016proposal, rocco2018a, jeon2018parn,seo2018attentive, PMD, WarpC} or a cost volume \cite{rocco2018b, DCCNet, pwarpc}. 
In this work, we propose a photo-sketch correspondence learning framework that explicitly estimates the dense flow field with image pair supervision.

%------------------------------------------------------------------------
\section{Conclusions}
\label{sec:conclusion}
What is needed to develop artificial systems that learn to perceive the visual world as keenly as humans do?
While artificial vision systems have made dramatic improvements in a variety of tasks, there remain key aspects of human image understanding that continue to pose major challenges.
Here we focused on one of these aspects: the ability to understand the semantic content of color photos and line drawings well enough to establish a detailed mapping between them.
Our paper introduces a new photo-sketch correspondence benchmark containing 150K human annotations of 6250 sketch-photo pairs across 125 object categories, augmenting existing photo-sketch benchmark datasets \cite{sangkloy2016sketchy}.
In addition, we conduct several experiments to evaluate a self-supervised approach to learning to predict these correspondences and compare this approach to several strong correspondence learning baselines.
Our results suggest that our approach combining contrastive learning and spatial transformer network is effective for capturing photo-sketch correspondences, but there remain systematic deviations from human judgments on the same task.
Taken together, we hope that these findings, along with our new fine-grained multimodal image understanding benchmark, will catalyze progress towards achieving more human-like vision systems.

\section*{Acknowledgements}
Many thanks to the members of the Cognitive Tools Lab and the Prof. Wang's Lab at UC San Diego for their helpful feedback and support. 
This work was supported by an NSF CAREER Award \#2047191 to J.E.F..
J.E.F is additionally supported by an ONR Science of Autonomy award and a Stanford Hoffman-Yee grant. Prof. Wang’s lab was supported, in part, by NSF CAREER Award IIS-2240014, DARPA LwLL, Amazon Research Award, and gifts from Qualcomm.

\nocite{langley00}

\bibliography{references}
\bibliographystyle{icml2023}

%%%%%%%%%%%%%%%%%%%%%%%%%%%%%%%%%%%%%%%%%%%%%%%%%%%%%%%%%%%%%%%%%%%%%%%%%%%%%%%
%%%%%%%%%%%%%%%%%%%%%%%%%%%%%%%%%%%%%%%%%%%%%%%%%%%%%%%%%%%%%%%%%%%%%%%%%%%%%%%
% APPENDIX
%%%%%%%%%%%%%%%%%%%%%%%%%%%%%%%%%%%%%%%%%%%%%%%%%%%%%%%%%%%%%%%%%%%%%%%%%%%%%%%
%%%%%%%%%%%%%%%%%%%%%%%%%%%%%%%%%%%%%%%%%%%%%%%%%%%%%%%%%%%%%%%%%%%%%%%%%%%%%%%
\clearpage
\appendix

\section{Details of the Photo-Sketch Correspondence Benchmark (\texttt{PSC6k})}
\label{sec:supp benchmark}

\subsection{Keypoint Sampling}
\begin{figure}[htp]
    \vspace{-1mm}
    \centering
    \includegraphics[width=0.5\textwidth, trim=2 4 0 2, clip]{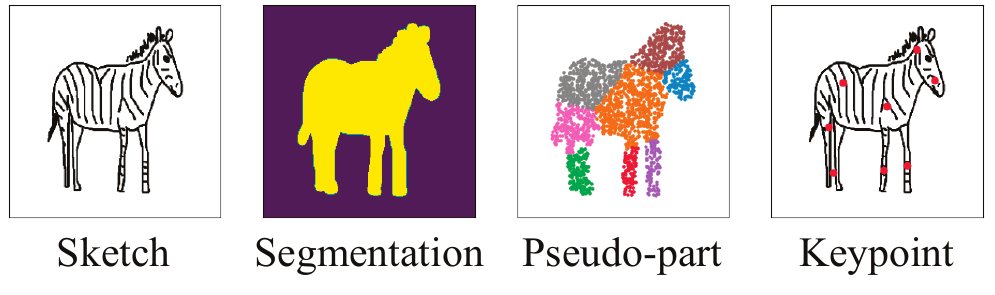}
    \vspace{-6mm}
    \caption{Example of the keypoint sampling process. We show the sketch, segmentation mask, pseudo-parts, and final keypoints.}
    \label{fig:keypoint sampling}
    \vspace{-5mm}
\end{figure}

We visualize the steps we take to sample eight keypoints spanning the object in \autoref{fig:keypoint sampling}. First, we fill in the outermost contour detected in the sketch to generate the segmentation of the object. In cases where multiple contours are detected due to unconnected strokes, we apply dilation and contour filling iteratively until all strokes are connected. We then cluster the pixels covered by the segmentation mask into 8 pseudo-parts, by building a nearest-neighbor-based affinity matrix over pixels and applying spectral clustering. Since the affinity between two pixels is defined by the shortest path instead of the L2 distance, it ensures a clustering that maintains the connectivity within each pseudo-part.

\subsection{Annotation Filtering}
In rare cases, for a given keypoint, one of the three annotations has an exceptionally large distance from the median location $\Tilde{x}$ of all three annotations, denoted as $d = \lVert x-\Tilde{x}\rVert_2^2$. We gather the distance $d$ for the 150,000 annotations that we collect and compute its mean and standard deviation. The annotations with $d$ of three standard deviations away from the mean are then considered outliers and excluded from the final determination of the centroids. This rejects 0.74\% of the annotations.

\section{Additional Evaluation on \texttt{PSC6k}}
\label{sec:supp eval}
\begin{table*}[htp]
\centering
\begin{tabular}{ll|c@{~~}c|c@{~~}c}
\toprule
&   & \multicolumn{2}{c}{\textbf{Transfer}} & \multicolumn{2}{c}{\textbf{Retrain}} \\
Sup & Methods & PCK-5 & PCK-10 & PCK-5 & PCK-10\\ \midrule
KP & HPF\cite{min2019hyperpixel} & 50.55 & 78.18 & -- & -- \\
 & CHM~\cite{min2021convolutional} & 40.52 & 69.91 & -- & --\\
 & PMD\cite{PMD}  & 28.62 & 63.95 & -- & --\\
 & CATs\cite{cho2021cats}  & \textbf{52.36} & \textbf{81.80} & --& --\\
\midrule
Pair & CNNGeo~\cite{rocco2018a}  & 27.59 & 57.71 & 19.19 & 42.57\\ 
 & WeakAlign~\cite{rocco2018a} & 35.65 & 68.76 & 43.55 & 78.60\\
 & NC-Net~\cite{rocco2018b} & 40.60 & 63.50 & --& -- \\
 & DCCNet~\cite{DCCNet} & 42.43 & 66.53 & -- &  -- \\
 & PMD~\cite{PMD} & 35.77 & 71.24 & -- & -- \\ 
 & WarpC-SemanticGLUNet~\cite{WarpC} & 48.79 & 71.43 & 56.78 & 79.70 \\ 
 & \textbf{Ours} (ResNet-18)  & -- & -- & 56.38 & \textbf{83.22}\\ 
 & \textbf{Ours} (ResNet-101)  & -- & -- & \textbf{58.00} & \textbf{84.93}\\ 
\bottomrule 
\end{tabular}
\caption{Comprehensive evaluation for photo-sketch correspondence learning.}
\label{tab:comprehensive comparison}
\end{table*}

\subsection{Methods with Stronger Supervision}
For a more comprehensive evaluation of existing correspondence learning methods on our \texttt{PSC6k} benchmark, we include methods with keypoint supervision in \autoref{tab:comprehensive comparison} and report their PCK for $\alpha = (0.05, 0.1)$. We report the performance of keypoint-supervised models in the transfer setting only (directly evaluate on the photo-sketch correspondence with pretrained weights), because they require supervision beyond what the sketchy training set provides. Interestingly, we observe that CATs \cite{cho2021cats} performs exceptionally well on the photo-sketch correspondence, even without retraining on photo-sketch pairs. This suggests its good ability of generalization.

\subsection{Training and Evaluation Details}
\textbf{Evaluation setting.} All methods are evaluated on our \texttt{PSC6k} benchmark using their original evaluation scripts. We make necessary edits to adapt the existing codes to \texttt{PSC6k}. 

\textbf{Training setting.} In the transfer setting, we use the pretrained weights on PF-Pascal provided by each method. In the retrain setting, we train the methods on the training split of the Sketchy dataset \cite{sangkloy2016sketchy} using their codes and default hyperparameters. Since there is no validation split, we do not select the best checkpoint and evaluate with the last checkpoint after training.

Since the training set of Sketchy dataset is 88X larger than that of PF-Pascal, it is impossible to keep the original training epochs and learning rate schedule in large models such as WarpC. Therefore, we make the following changes to the series: instead of training 100 epochs as in the original settings, we find that training for 2 epochs on Sketchy has already guaranteed an optimal performance (since it leads to $1.77\times$ iterations compared to the original training scheme). We reduce the learning rate to $0.125\times$ in the second epoch to approximate the original LR schedule of the method.

\textbf{Causes of blank entries.} The retrain performance of several methods are left blank for the following reasons: 
\begin{itemize}
    \itemsep0em
    \item The method requires stronger supervision than what the Sketchy training set provides. This applies to all methods with keypoint supervision. 
    \item The method fails to converge on the photo-sketch correspondence task, which is observed in NC-Net and DCCNet. We hypothesize that since the sketch samples are out-of-domain, their cost volume optimization blocks fail to handle the large disparity between the representations of photos and sketches.
    \item The method does not provide codes for training: PMD.
    \item In addition, methods that did not release source code, failed to execute, or did not provide pre-trained weights are excluded from the table.
\end{itemize}

\section{Generalization to Unseen Categories}
\label{sec:supp unseen}
To analyze the generalization capability of our proposed model, we evaluate its performance on categories that were not included during the training phase. Specifically, we randomly sample N categories from the full set of 125 categories in the Sketchy dataset, and hold them out during the training of both the feature encoder and warp estimator. Then we evaluate the model performance of correspondence estimation on these N held-out categories. We conduct experiments for N=10 and N=20. The mean performance and standard deviation were calculated based on three randomly sampled held-out splits for each of the two conditions. The results are presented in \autoref{tab:unseen}.

As shown in the table, our method maintains a very decent performance on the 10/20 categories absent during training, with a decrease of -0.33\%/-0.26\% for 10 held-out categories and a decrease of -0.50\%/-0.37\% for 20 held-out categories. This shows that our method is robust in generalization to unseen categories.

\begin{table}[hp]
\centering
\resizebox{0.45\textwidth}{!}{
\begin{tabular}{l|c@{~~}c}

\# Categories (N) & PCK-5 ($\pm$ std) & PCK-10 ($\pm$ std)\\ \midrule
0 & 56.38 & 83.22 \\
10 & 56.05 (0.20) & 82.96 (0.15)\\
20 & 55.88 (0.27) & 82.85 (0.18)\\
\bottomrule
\end{tabular}
}
\caption{Model performance on unseen categories.}

\label{tab:unseen}
\end{table}

\section{Additional Qualitative Results}
\label{sec: supp qualitative}
\begin{figure}[ht]
    \centering
    \includegraphics[width=0.5\textwidth, trim=0 0 0 0, clip]{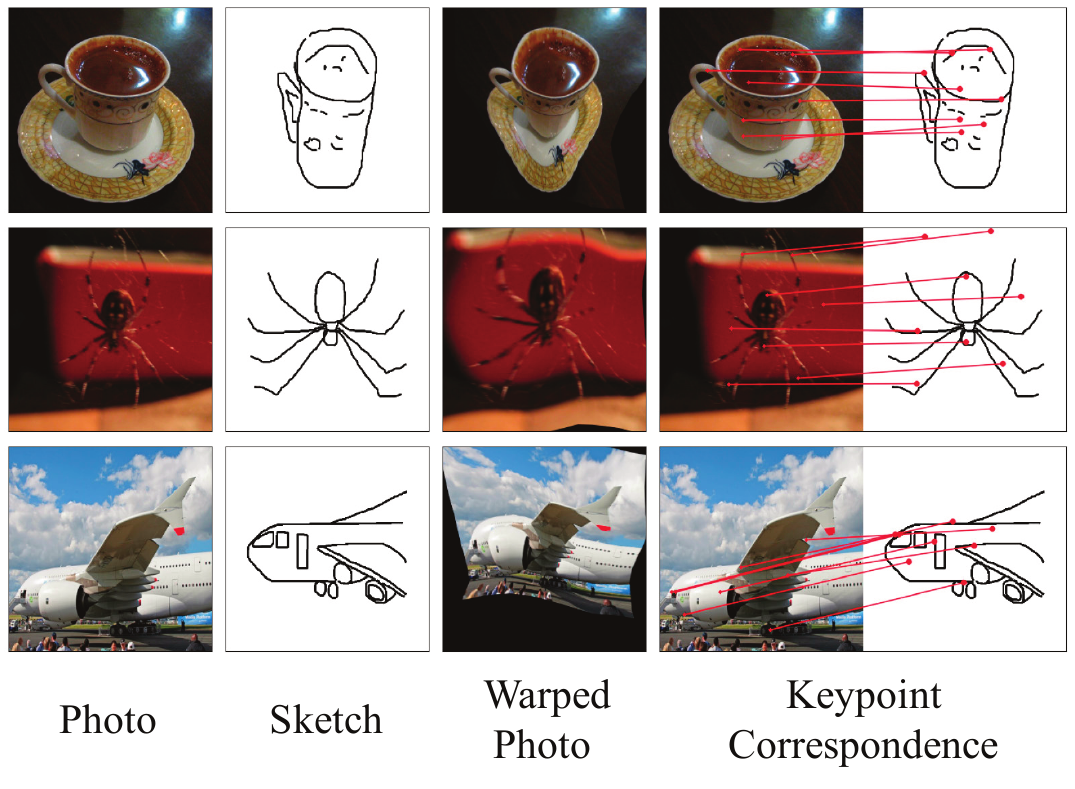}
    \vspace{-6mm}
    \caption{Examples of three typical failure patterns. The method has worse performance for: 1) commonly co-occurred objects, 2) fine structures, and 3) non-continuous transformations.}
    \label{fig:failure pattern}
    \vspace{63mm}
\end{figure}
We show typical failure patterns in \autoref{fig:failure pattern}. Specifically, the model has degraded performance in 1) discriminating between commonly cooccurred objects; 2) aligning fine structures due to low-resolution feature maps; and 3) handling non-continuous transformation caused by large changes in perspective and structure, which violates the continuity assumption in warp-based models that close points should be mapped to close locations. We believe that they are the main problems that need to be addressed in future studies.

\begin{figure*}[ht]
    \centering
    \includegraphics[width=0.95\textwidth, trim=5 5 5 5, clip]{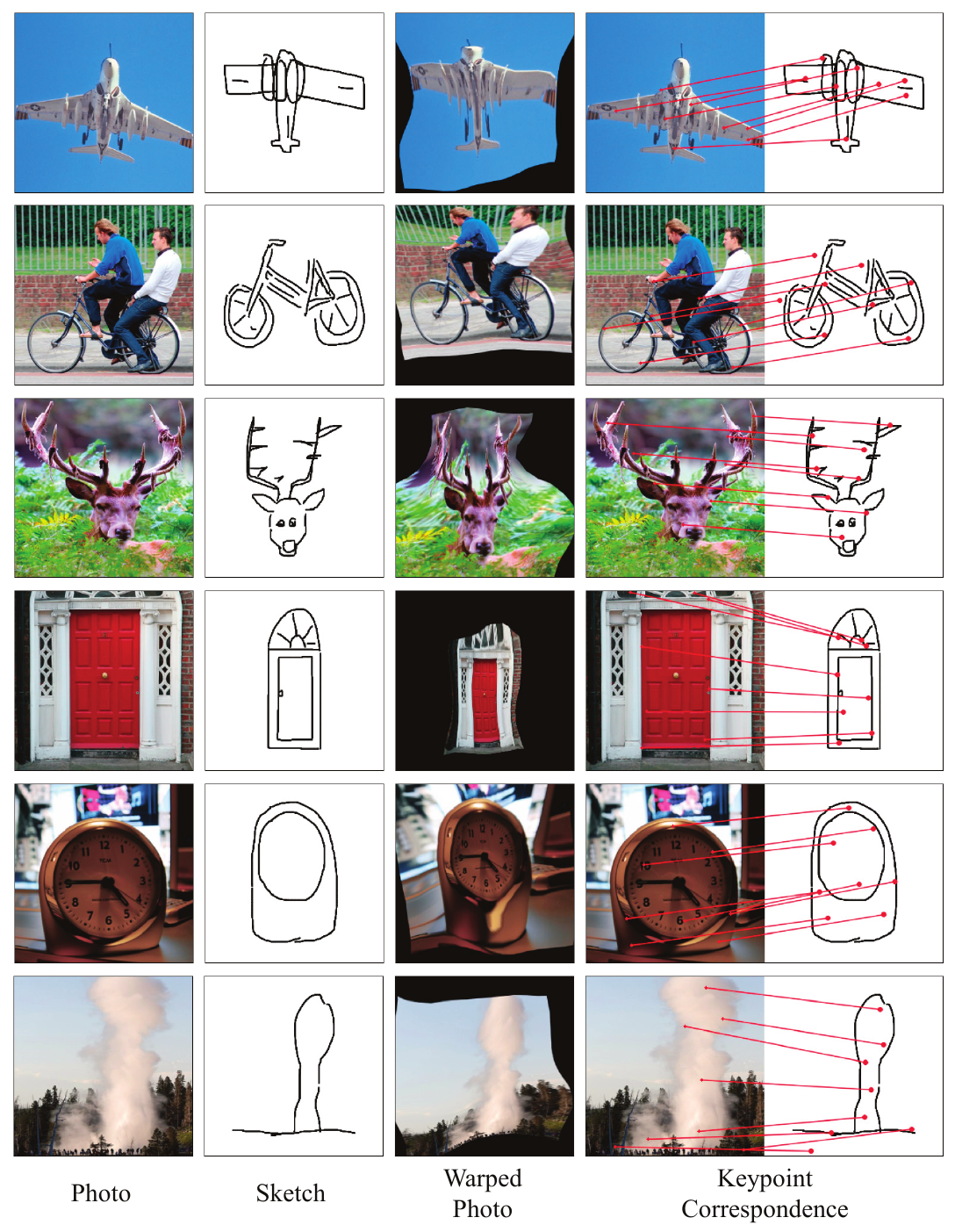}
    \caption{More alignment examples on the \texttt{PSC6k} dataset.}
    \label{fig:visualization1}
    \vspace{-2mm}
\end{figure*}

\begin{figure*}[ht]
    \centering
    \includegraphics[width=0.95\textwidth, trim=5 5 5 5, clip]{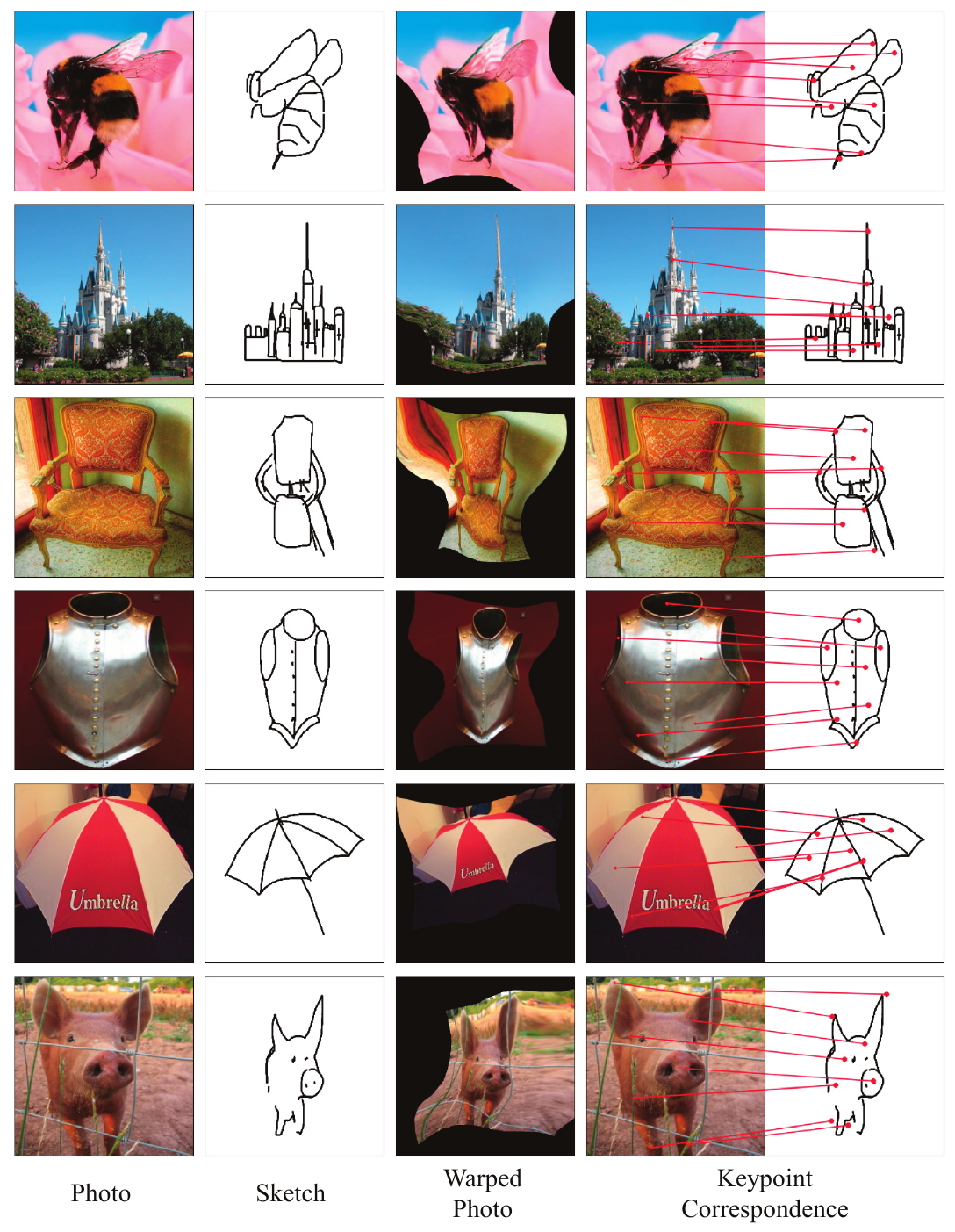}
    \caption{More alignment examples on the \texttt{PSC6k} dataset.}
    \label{fig:visualization2}
    \vspace{-2mm}
\end{figure*}

\begin{figure*}[ht]
    \centering
    \includegraphics[width=0.95\textwidth, trim=5 5 5 5, clip]{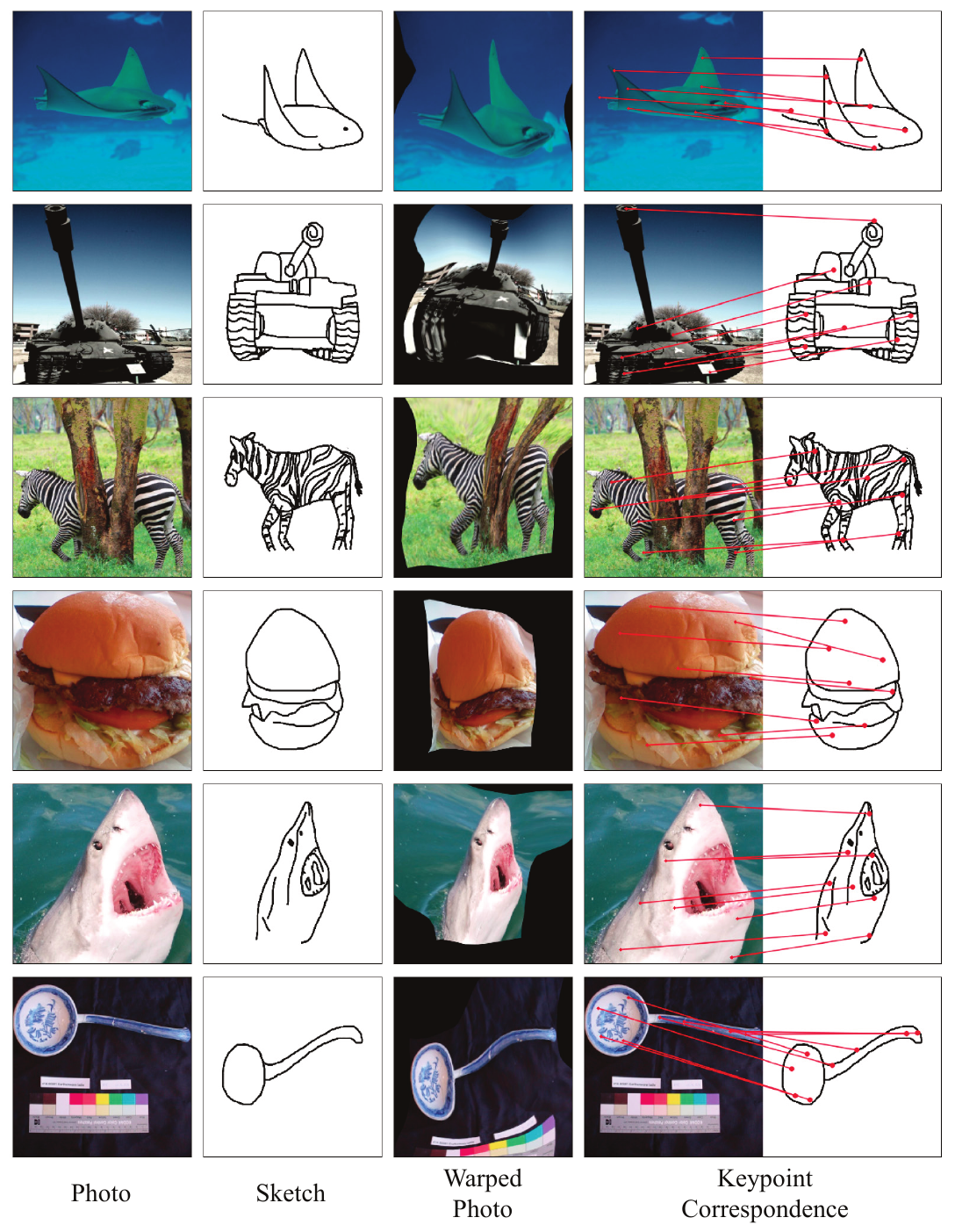}
    \caption{More alignment examples on the \texttt{PSC6k} dataset.}
    \label{fig:visualization3}
    \vspace{-2mm}
\end{figure*}

Lastly, we exhibit more examples of photo-sketch correspondence predicted by our model (\autoref{fig:visualization1}, \autoref{fig:visualization2}, \autoref{fig:visualization3}).

%%%%%%%%%%%%%%%%%%%%%%%%%%%%%%%%%%%%%%%%%%%%%%%%%%%%%%%%%%%%%%%%%%%%%%%%%%%%%%%
%%%%%%%%%%%%%%%%%%%%%%%%%%%%%%%%%%%%%%%%%%%%%%%%%%%%%%%%%%%%%%%%%%%%%%%%%%%%%%%

\end{document}